\documentclass{article}

\usepackage{amsmath,amsfonts,amssymb,times,graphicx,natbib,algorithm,algorithmic,hyperref}

\usepackage[accepted]{whi2016}

\icmltitlerunning{Interpreting extracted rules from ensemble of trees: 
	Application to computer-aided diagnosis of breast MRI}

\begin{document}
\twocolumn[
\icmltitle{Interpreting extracted rules from ensemble of trees: \\
	Application to computer-aided diagnosis of breast MRI}

\icmlauthor{Cristina Gallego-Ortiz}{cgallego@sri.utoronto.ca}
\icmlauthor{Anne L. Martel}{anne.martel@sri.utoronto.ca}
\icmladdress{Department of Medical Biophysics, University of Toronto}
\icmladdress{Department of Imaging Research, Sunnybrook Research Institute, Toronto}

\vskip 0.3in
]

\begin{abstract}
High predictive performance and ease of use and interpretability are important requirements for the applicability of a computer-aided diagnosis (CAD) to human reading studies. We propose a CAD system specifically designed to be more comprehensible to the radiologist reviewing screening breast MRI studies. Multiparametric imaging features are combined to produce a CAD system for differentiating cancerous and non-cancerous lesions. The complete system uses a rule-extraction algorithm to present lesion classification results in an easy to understand graph visualization.
\end{abstract}

\section{Introduction}
\label{sec:intro}

Breast MRI is the imaging modality with the highest sensitivity for detecting breast cancer \citep{Chiarelli2014}. Currently, it is recommended for screening high risk women and its use in screening average risk women remains a debate \citep{Kuhl2010}. Computer-aided diagnosis (CAD) has been proposed as a tool to aid the diagnostic decision process in breast MRI. Current CAD algorithms typically use supervised learning to learn discriminative imaging patterns between cancers and non-cancers. An important criterion in CAD is the interpretability of the models inferred from images and how they may contribute to the human expert making a clinical decision \citep{Jorritsma2015}. Such models can play a key role in reducing overdiagnosis while maintaining a high cancer yield, if they are able to downgrade accurately the suspicion of a lesion on the basis of imaging findings. 

Recently, in breast MRI CAD research, a lot of attention has been devoted to translating knowledge from clinical diagnostic practice into suitable CAD features and algorithms \citep{Gubern-Merida2014,Gallego-Ortiz2015}, but less interest has been granted to the interpretability of CAD systems.

Our approach to increase interpretability in CAD is based on two main strategies: First, by designing computer-extracted features that reflect simple and interpretable lesion characteristics and second, by summarizing CAD results in the form of comprehensible rules that can be explained in terms of the lesion characteristics. 

During breast MRI examination, a radiologist estimates the level of suspicion for malignancy by assessing signal enhancement on T1-weighted (T1w) dynamic contrast enhanced (DCE) MRI (i.e. how tissues brightness changes over time) as well as the morphology of a lesion. A typical malignant lesion will exhibit heterogeneous signal enhancement, and irregular shape with spiculated margins, while a typical benign lesion has more round shape, frequently with smooth margins. Computer-extracted features are typically engineered to reflect these diagnostic criteria. For example, circularity and margin irregularity are morphologic features and signal enhancement characteristics are measured with kinetic and texture features. On T2-weighted (T2w) imaging, radiologists report the signal intensity (SI) based on categories such as Hypointensity or hyperintensity or using a relative signal intensity metric, such as the ratio of lesion to muscle intensity as a tissue reference. When reported, T2w SI reflects relevant diagnostic information \citep{Kuhl1999b,Moran2014}. In this sense, features designed to reproduce radiologists' criteria on T2w imaging enriches CAD knowledge, when these features are added to image-only features. 

Overall, we present a CAD system specifically designed to be more comprehensible to the radiologist reviewing screening breast MRI studies. T1w and T2w features are combined to produce a multiparametric CAD system for differentiating cancerous and non-cancerous lesions. The complete CAD system uses a rule-extraction algorithm to present lesion classification results in an easy to understand graph visualization.

\section{Materials}
\label{sec:materials}

We retrospectively identified $435$ women aged $48.75 \pm 10.6$ years (mean $\pm$ std), with $627$ breast lesions reported in imaging studies performed prior to biopsy in which ground truth histopathology was obtained. Only patients without prior breast surgery or cancer radiotherapy/chemotherapy treatment were considered in this study. Lesions included $245$ malignant (cancerous) and 382 benign (non-cancerous) lesions.  

The imaging protocol consists of two sagittal T2-weighted FSE fat saturated acquisitions, one for each breast, and bilateral sagittal DCE T1-weighted fat saturated 3D FSPGR images. Dynamic protocol includes one pre-contrast and 4 post-contrast acquisitions. DCE-MRI volumes have less than 1 mm in-plane resolution, 2 to 3 mm slice thickness, and under 120 sec/dynamic acquisition. Use of breast MRI datasets for CAD research was approved by our institutional review board, waiving informed consent. 

\section{Methods}
\label{sec:methods}

\subsection{Lesion segmentation and features}
Lesions were segmented as Volume-of-interest (VOI) based on the area of enhancement on T1w CE-MRI using the location quadrant and clock position from the radiological report. An interactive seeded 3D region growing algorithm was employed to segment the lesion VOI. 
A total of 99 T1w CE-MRI derived features comprising 34 kinetic, 19 morphologic and 46 texture features were used, with additional 45 T2w features. 80 features based on the individual post-contrast intensities (20 per post-contrast volume) and 20 based on the dispersion of signal enhancement spatial distribution\footnote{Dispersion measures the distribution and clustering of enhancements at each of the post-contrast time points}.

By overlaying the segmented lesion VOI on the T2w volume, we extracted from the corresponding area of enhancement, the T2w signal intensity (SI) and the first 4 statistical moments (i.e. mean, variance, skewness, and kurtosis). Morphological features such as margin sharpness were extracted from the T2w VOI. 

\subsection{T2w features derived using predictive models}
Two interpretable T2w CAD features are proposed: 1) the T2w signal intensity category according to BIRADS lexicon \citep{ACR-BIRADS} and the 2) ratio of signal intensity of the lesion to the pectoralis muscle as tissue of reference (LMSIR). These T2w CAD features are derived using predictive models that use raw imaging data as exploratory features. Using a rectangular 5 $mm^3$ ROI, placed manually in the pectoralis muscle area, we extracted T2w reference signal intensity of the muscle.

A predictive model of BIRADS T2w signal intensity was developed using ensembles of boosting classification trees. The proposed predictive model uses image-based features as exploratory variables and the radiologists' reported BIRADS category as the target variable. The predictor consists of a sequence of 3-steps of classification predicting incremental categories of T2w signal intensity: step-1) No signal reported v.s signal reported, step-2) Hypointense v.s Hyperintense, and step-3) slightly Hyperintense v.s Hyperintense. 

Similarly for the LMSIR, a numerical predictor based on ensembles of boosting regression trees is developed. As proposed in \citep{Ballesio2009c}, LMSIR was measured in the training population as the signal intensity average in the T2w VOI divided by the signal intensity average in the muscle VOI. This measured LMSIR is then used as the target variable and image-based features as exploratory variables.

\subsection{Rule-extraction from the CAD lesion classifier}
The proposed CAD system for lesion classification uses a rule-extraction algorithm that summarizes classification results, based on the Node harvest algorithm \cite{Meinshausen2010}. The main idea with Node harvest is to add interpretability to a random forest classifier by translating node split functions to interpretable classification rules.

Given paired observations $\{(x_1, y_1),...,(x_n,y_n)\}$, where each $x_i$ represents an input feature and $y_i$ the corresponding class label, a classification rule is a procedure in which we assigned individual observations to classes. From any tree in the ensemble, an internal node $Q_j$ is associated with a binary split or test function over the inputs $h_{Q_j}(X, \theta_j)$. In turn, each leaf node is associated with a predictor model that assigns a conditional class probability given the input features $P(y=1|X)$. An internal node test function can be expressed as a classification rule $\{C \Rightarrow T \}$, where $C$, is referred to as the condition of the rule, and $T$ is the outcome of the rule. With node harvest, the conditions of a rule can be explained in terms of the internal node functions $h_{Q_j}(X, \theta_j)$ and the outcome of the rule in terms of mean response at a node $\mu_{Q_j}$. Since $\mu_{Q_j}$ is the fraction of cases of the positive class at the node, rule predictions are naturally in the interval [0-1].
So rules extracted with Node Harvest can be expressed as $\{h_{Q_j}(X, \theta_j) \Rightarrow P(y=1|x_i \in Q_j) \}$.

\subsection{Interpreting prediction outcomes in terms of simple classification rules}
Node Harvest starts by building a random Forest \cite{Breiman2001} with the input features. Then it proceeds to find the best subset of classification rules of all the possible rules that can extracted from the forest. The best subset is found with an optimization over the selected subset so that when combined by weighted voting, they produce the lowest classification error for a given set of observations. 

CAD lesion classification prediction for a new test case is the weighted average prediction:
$$  P(y=1|x_{test}) = \frac{\sum_{j \in G_x} w_j \mu_j }{\sum_{j \in G_x} w_j} $$

Where $ G_x \equiv \{ j: x \in Q_j \}$ is the collection of nodes where the observation $x_{test}$ falls into. The set $G_x$ is always non-empty since the root node is always included in the set. Additionally, $G_x$ can always be decomposed into its constituent rules, so that prediction outcomes can be explained as combinations of all the classification rules the observation meets.

\subsection{Visualization of CAD lesion classification}
\begin{figure*}
\vskip 0.2in
\begin{center}
\centerline{\includegraphics[width=2\columnwidth]{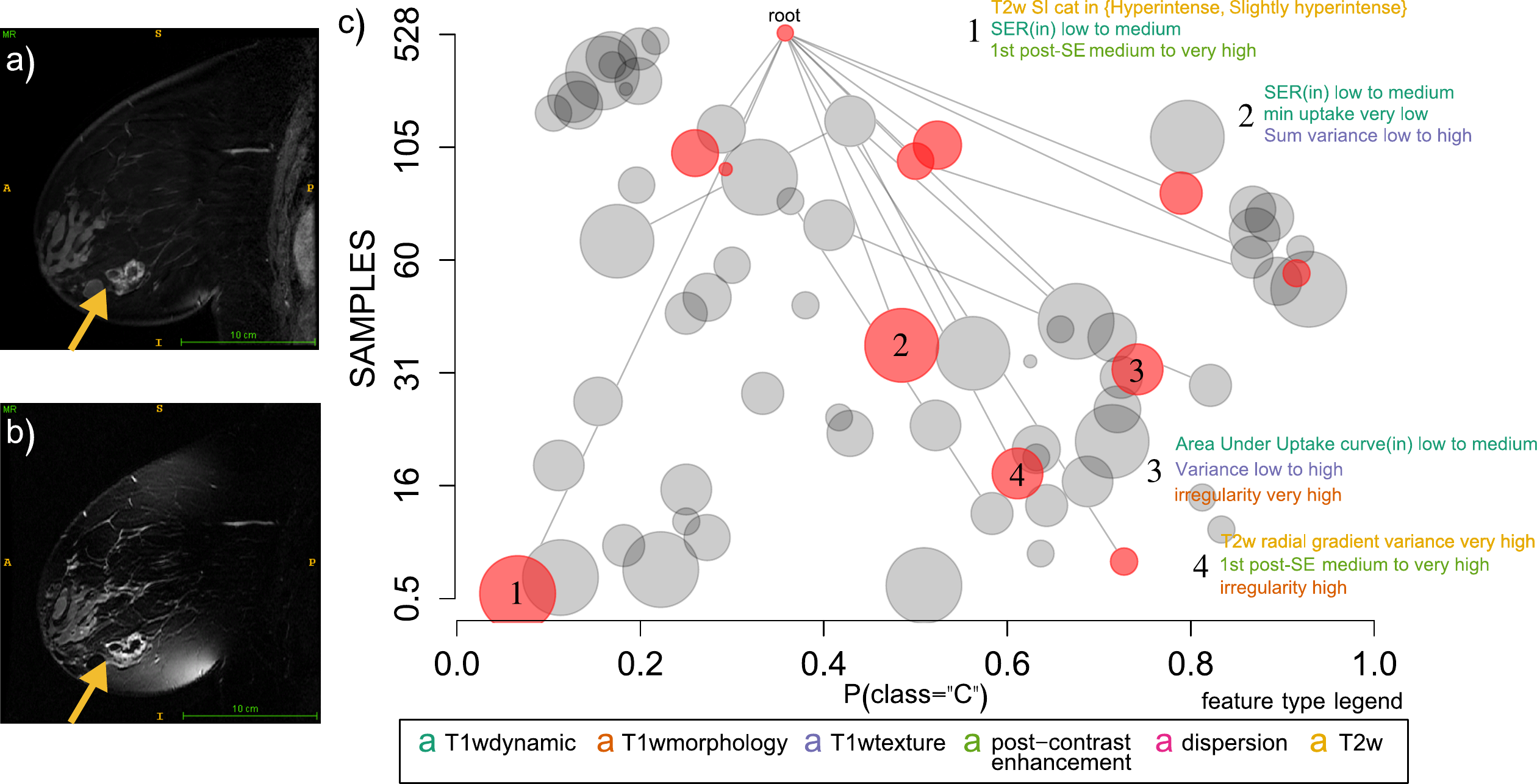}}
\caption{Illustrative CAD diagnosis of a benign lesion (Fibroadenoma) a) 1st post-contrast T1w. b) T2w FSE slice through the lesion. c) CAD classifier output.}
\label{fig:Fig1}
\end{center}
\vskip -0.2in
\end{figure*}

\begin{figure*}
\vskip 0.2in
\begin{center}
\centerline{\includegraphics[width=2\columnwidth]{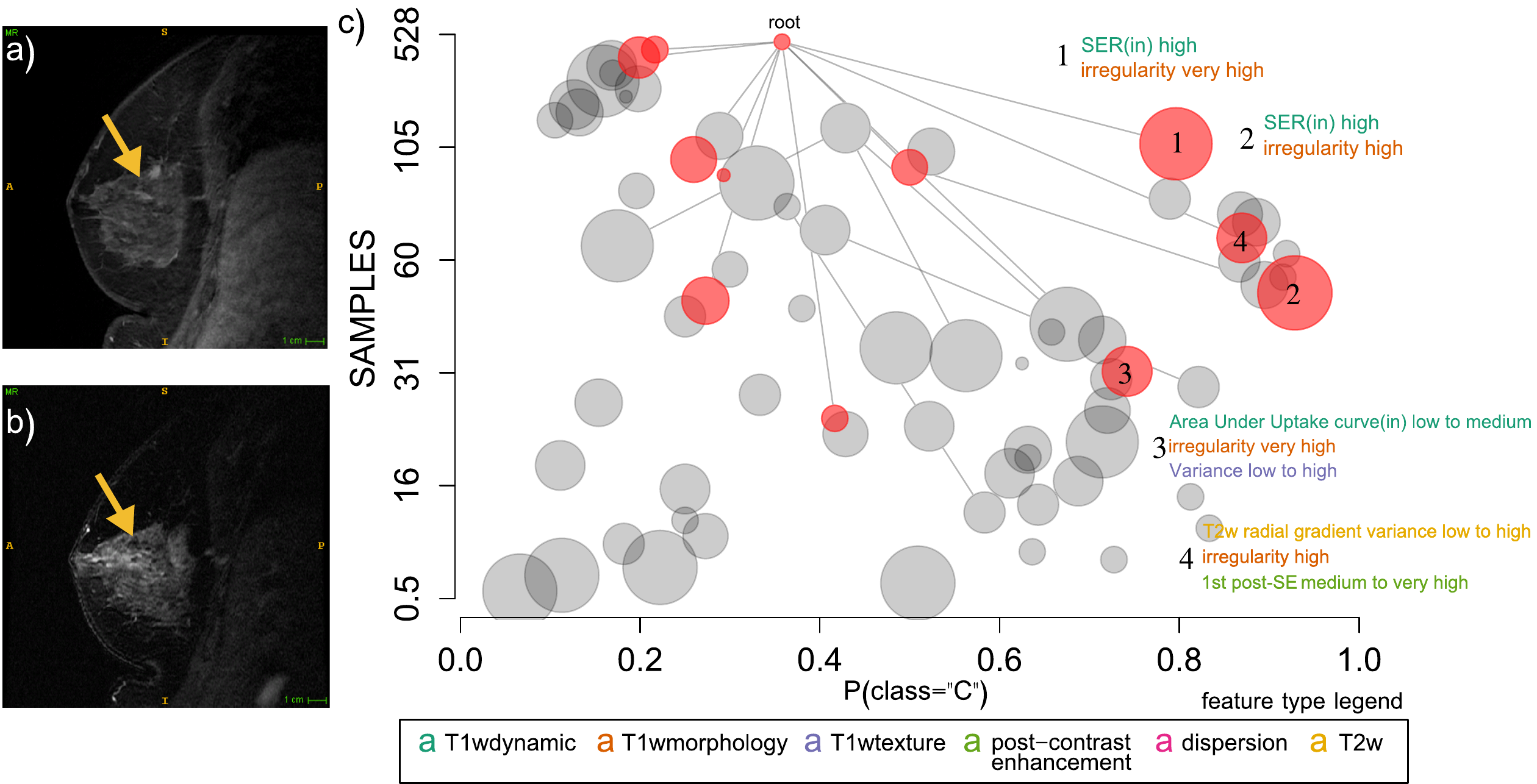}}
\caption{Illustrative CAD diagnosis of a malignant lesion (insitu-ductal carcinoma) a) 1st post-contrast T1w. b) T2w FSE slice through the lesion. c) CAD classifier output.}
\label{fig:Fig2}
\end{center}
\vskip -0.2in
\end{figure*}

The most interesting property of classification rules extracted with node harvest is that they are graphically interpretable. A prediction on a new test case can be visualized in a tree-like graph (see Figures \ref{fig:Fig1} and \ref{fig:Fig2}). To interpret this tree: start at the root node, and then proceed to only the nodes where the test observation falls into. Each of these nodes produce a class prediction for the new test case. Graphically, the predicted class is the weighted mean across the x-axis of the highlighted nodes. In addition, nodes are plotted so that their size is proportional to their weight. The y-axis represents the node sample size.

To make our CAD lesion classification more interpretable, we categorized node conditions with thresholds. A node condition threshold for a numerical feature can be mapped to corresponding intervals with equal probabilities according to its univariate distribution. We transform feature quartiles: 0\%(min), 25\%(Q1), 50\%(Median or Q2), 75\%(Q3) and 100\%(max) to ordinal categories: very low, low, medium, high and very high. Therefore, binary decision thresholds at a node can be transformed into ordinal categories. We show the graphical interpretation of some test case predictions for illustration in the results section.

\section{Results}
For T2w BIRADS categories, the cascade-based predictor produced median ROC AUCs of 0.71, 0.82, 0.85 for stage 1, 2, and 3 respectively, during 10-fold cross validation. The predicted category with the highest agreement with the radiologist reported T2w BIRADS category was Slightly Hyperintense (86\% accuracy) followed by Hypointense or not seen (79\% accuracy), and Hyperintense with (0.76\% accuracy). The “None” signal or not reported category was reproduced with 56\% accuracy.

Measured LMSIR from ROIs placed in the muscle region ranged from 0.22 to 26.8 with a median value of 2.85. The median root mean square error (RMSE) in the predicted LMSIR during 10 fold cross-validation was 0.89 IQR: [0.47-1.58]. RMSE range was [0.06-20.6]. The median AUC ROC performance for the node harvest lesion classifier was 0.84 95\% CI (0.81-0.86) based on all 627 pooled lesion classification results from 10 fold cross-validation. 

\subsection{Illustration: Aided-interpretation of CAD analysis}
Figures \ref{fig:Fig1} and \ref{fig:Fig2} illustrate the use of the proposed interpretable CAD classifier. All nodes are plotted but only the nodes where the test observation falls into are highlighted. Only the rules from the 4 most relevant nodes or the nodes with the 4 largest weights are indicated by a numeral, since classification rules derived from these nodes heavily influence the probability estimate of lesion malignancy. Classification rules are shown, and their conditions mapped to the corresponding feature ordinal categories (i.e. low, medium or high value). In addition, features are color coded according to their feature group. Panel a) shows the appearance of the lesion in the 1st post contrast T1w image and b) the T2w lesion appearance. Panel c) shows the node harvest classifier output visualized as a graph.

Figure~\ref{fig:Fig1} illustrates a Fibroadenoma, presenting as a multilobulated rim enhancing lesion with rapid initial enhancement with plateau on T1w CE-MRI.  On T2w imaging, the center appears hypointense and the rim that is enhancing hyperintense. Predicted BIRADS category was Hyperintense and was in agreement with radiological report. The lesion classifier produced a probability of malignancy of 0.45. Since the CAD determines that a lesion is benign if $P(class=”C”) < 0.5$, the classifier achieves a correct lesion classification. The predicted T2w BIRADS appears in the node with the largest weight (Node 1 with weight of 22\% and $P(class=”C) = 0.07$. 

Figure~\ref{fig:Fig2} illustrates an \emph{in-situ} ductal carcinoma, presenting as lesion with moderate initial and persistent enhancement  on T1w CE-MRI. The T2w imaging appearance was not reported. The lesion classifier produced a probability of malignancy of 0.64. Since the CAD determines that a lesion is malignant if $P(class=”C”) \geq 0.5$, the classifier achieves a correct lesion classification. The two largest nodes show a combination of T1w dynamic, T1w morphology and T1w 3D texture features and all produce a $P(class=”C)$ higher than 0.5. 

\section{Conclusions} 
\label{sec:conclusion}
High predictive performance, but also ease of use and interpretability are important requirements for the applicability of a computer-aided diagnosis (CAD) to human reading studies. We strived to take into consideration CAD interpretability when designing our CAD system in this study. 

\section*{Acknowledgements}
This research is funded by the Canadian Breast Cancer Foundation (CBCF) - Ontario Region, and the Ontario Institute for Cancer Research (OICR) through funding provided by the Government of Ontario. The authors thank Sharmila Balasingham, BSc, for help curating patient clinical information from multiple databases and Nim Li, BComm, for technical assistance with the research database. 

\bibliography{references}
\bibliographystyle{icml2016}

\end{document}